\documentclass{article}

\usepackage{PRIMEarxiv}

\usepackage[utf8]{inputenc} 
\usepackage[T1]{fontenc}    
\usepackage{hyperref}       
\usepackage{url}            
\usepackage{booktabs}       
\usepackage{amsmath,mathdots}        
\usepackage{amssymb}
\usepackage{nicefrac}       
\usepackage{microtype}      
\usepackage{lipsum}
\usepackage{fancyhdr}       
\usepackage{graphicx}       
\graphicspath{{media/}}     

\title{From Smart Sensing to Consciousness: An info-structural model of computational consciousness for non-interacting agents}

\author{
  Gerardo Iovane \\
  Department of Computer Science \\
  University of Salerno \\
  Fisciano, Italy \\
  \texttt{giovane@unisa.it} \\
   \And
  Riccardo Emanuele Landi \\
  Rigenera S.r.l. \\
  Rome, Italy \\
  \texttt{riccardo.landi@rigenera2020.it}
}

\begin{document}
\maketitle

\begin{abstract}
This study proposes a model of computational consciousness for non-interacting agents. The phenomenon of interest was assumed as sequentially dependent on the cognitive tasks of sensation, perception, emotion, affection, attention, awareness, and consciousness. Starting from the Smart Sensing prodromal study, the cognitive layers associated with the processes of attention, awareness, and consciousness were formally defined and tested together with the other processes concerning sensation, perception, emotion, and affection. The output of the model consists of an index that synthesizes the energetic and entropic contributions of consciousness from a computationally moral perspective. Attention was modeled through a bottom-up approach, while awareness and consciousness by distinguishing environment from subjective cognitive processes. By testing the solution on visual stimuli eliciting the emotions of happiness, anger, fear, surprise, contempt, sadness, disgust, and the neutral state, it was found that the proposed model is concordant with the scientific evidence concerning covert attention. Comparable results were also obtained regarding studies investigating awareness as a consequence of visual stimuli repetition, as well as those investigating moral judgments to visual stimuli eliciting disgust and sadness. The solution represents a novel approach for defining computational consciousness through artificial emotional activity and morality.
\end{abstract}

\keywords{artificial consciousness \and attention \and awareness \and consciousness \and decision support systems \and cognitive systems \and artificial intelligence}

\section{Introduction}
\label{sec:introduction}

Artificial Consciousness (AC) is a branch of Artificial Intelligence that aims to build computational models inspired by the functioning of human consciousness. In particular, rather than artificially reproducing the above phenomenon, the goal is to constitute, or at least identify, consciousness-like processing to the machine. The related field of research is mainly associated with computer science, but the modeling of human cognition permits enhancing the multidisciplinarity of the results. AC is particularly important for designing novel methodologies and algorithms in the field of Artificial Intelligence since the study of human cognition is significant for defining increasingly sophisticated cognitive systems. Social robots or virtual agents need to reach suitable flexibility in interacting with human beings, both from an emotional and behavioral point of view. In the context of human-machine interaction, the deployment of software solutions which guarantee perceptual and emotional adaptation to the environment allows achieving significant results in terms of user satisfaction \cite{ritschel2017real, paiva2017empathy, bonarini2020robot}.

The phenomenon of consciousness has never been defined precisely. Some scholars distinguish consciousness from awareness, while others conceive the two concepts as synonyms; others, instead, describe the phenomenon in terms of perception, emotion, and morality. Further difficulties arise from the fact that only a part of consciousness seems to result in an objective functioning; most of what is conceived as “conscious experience” appear somehow subjective, i.e., dependent on personal judgments, feelings, or comparisons with others. The above observation requires constructing computational models of consciousness that possibly lack generality since there seems to be the necessity of introducing the element of subjectivity in the machine. The reproduction of the \textit{good and evil} conception, i.e., the translation of morality to the machine, represents one of the closest horizons towards which Artificial Intelligence is planning to go beyond, but the vision of AC as “artificial morality,” to our knowledge, has not been explored sufficiently. This is probably because research has been focused on proving the objective, rather than the subjective, functioning of consciousness. Furthermore, while a lot of studies focus on the correlates between brain and consciousness, no suitable investigation has been performed concerning the correlates between emotions and consciousness. For this reason, the present study was intended to provide an interpretation for studying consciousness, rather than just proposing a computational model. There is no intention, in any way, of trying to demonstrate the functioning of consciousness; the present study should be considered as one of the admissible models which reductively explain certain mechanisms of the phenomenon of interest. As it can be verified in the experimental Section of the present work, certain evidence corresponded with the analyzes performed on human beings, while others did not. 

The present study focuses on the definition and experimentation of a computational model of consciousness for non-interacting agents, i.e., for agents which merely observe, without carrying out actions towards the environment, with a particular focus on subjectivity. It was decided to develop a model considering human cognitive tasks as a hierarchy of layers involving the concepts of sensation, perception, emotion, affection, attention, awareness, and consciousness. The model was founded on a previously conducted work \cite{iovane2020smart}, in which the authors indicated the research path and modeled the first four cognitive layers (i.e., those associated with the concepts of sensation, perception, emotion, and affection). Re-executing the prodromal solution, it was decided to model the cognitive layers associated with the concepts of attention, awareness, and consciousness. The proposed model of computational consciousness was defined with reference on the following assumptions: i) attention depends on sensations, perceptions, emotions, and affections, and acts as a modulating factor for awareness to happen (i.e., without a suitable layer of attention, no awareness can happen); ii) awareness is not the same concept as consciousness, since the former is implied in acquiring experiences, while the latter in connecting and associating experiences with moral semantics; iii) consciousness depends on the awareness and associates experiences with semantics through personal and social morality, which in turn depends on the emotional activity. 

The study is structured as follows. Section \ref{sec:consciousness_background} presents a theoretical background on the studies concerning the phenomenon of consciousness, highlighting which, to our opinion, are the most important studies, definitions, and horizons in the research scenario, as well as the key concepts deployed for commenting on our modeling choices and experimental results. In Section \ref{sec:related_work} the latest advancements in AC are mentioned and the context in which it was intended to provide our contributions is indicated, while in Section \ref{sec:prodromal_study} the prodromal study based on which the proposed model of computational consciousness was developed and experimented is described. Section \ref{sec:proposed_model} is reserved for the formal definition of the proposed solution concerning the cognitive tasks of attention, awareness, and consciousness, while Section \ref{sec:experimental_results} describes the results obtained by experimenting the solution through a dataset of visual stimuli. Finally, in Section \ref{sec:discussion} the achieved results are discussed and compared with the scientific evidence in the studies of consciousness, while in Section \ref{sec:conclusions} the conclusions are drawn and future work is indicated.

\section{Consciousness: a theoretical background}
\label{sec:consciousness_background}

Consciousness is a complex phenomenon addressed for years, from a multidisciplinary point of view, by many branches of research, such as, e.g., Phenomenology, Philosophy of mind, Neuroscience, and Psychology. An exhaustive synthesis of the scientific debate on the subject of consciousness is difficult to provide and is out of the purpose of the present study. The dissertation is limited to mentioning the theories that, to our knowledge, enhance the most important aspects of the phenomenon. Further hints on scientific evidence are supplied in Sections \ref{sec:proposed_model} and \ref{sec:discussion}, in which the proposed model of computational consciousness is defined and its results are compared with those identified by further studies involving human beings, respectively.

Consciousness is often dyadically declined as \textit{objective} and \textit{subjective}: the first indicates states of awareness related to objective stimuli, i.e., instances of the reality acquired consciously and not affected by subjective factors, such as judgments or feelings; the second may be intended as awareness of objective stimuli from a subjective point of view, e.g., associating perceived stimuli to personal judgments or feelings. A different categorization is that concerning \textit{phenomenal} and \textit{access} consciousness \cite{block1995confusion}: the first refers to the cognition reserved for raw experiences of the body, such as sensations, movements, and emotions, also called \textit{qualia}; the second indicates the processing of information acquired from the environment through language, reasoning, and personal evaluation.

In Neuroscience, studies concerning the correlates of brain activity with a subject's reported experiences, also called the \textit{neural correlates of consciousness}, play a fundamental role. The aforementioned investigations mainly employ EEG (Electroencephalography) and fMRI (Functional Magnetic Resonance Imaging) to analyze the subject's brain activity. The correlation of physical and mental processes, which investigation was often called “the hard problem of consciousness,” \cite{chalmers1996conscious} has not been proved yet. Neuroscience tries to explain the functioning of consciousness in terms of neuronal effects, but exhaustive evidence is still missing; for instance, the issue of \textit{binding} neural activity to the experience of consciousness has not been solved yet \cite{singer2001consciousness}. Graziano and Kastner \cite{graziano2011human} supported the thesis for which consciousness would function through the mechanisms of social perception. They consider the phenomenon of consciousness as the perception of the awareness of an external subject to a stimulus; this process, rather than taking place outside, would take place inside the mind. However, the correlation between consciousness and the areas of the brain associated with social perception considered by the authors, i.e., superior temporal polysensory area and temporoparietal junction, is still debated as an exhaustive explanation of the phenomenon. Other discussed theories are those of the \textit{holonomic brain} \cite{pribram1986holonomic} and \textit{Orch-OR} \cite{hameroff2014consciousness} theories, which try to explain consciousness in terms of quantum neuronal effects.

A theory for describing consciousness in terms of quantity and quality was provided by Tononi \cite{tononi2004information}. With the Information Integration Theory, the author proposed to describe consciousness as the capacity of a system to integrate information. In particular, the author defined the quantity of consciousness as the \textit{information integration} $\phi$, i.e., the effective information of the minimum information bipartition of a complex. In this context, the complex represents a subset characterized by $\phi > 0$ that is not part of a subset characterized by higher $\phi$. The quality of consciousness, instead, was modeled as the informational relationships among the elements of a complex, i.e., the related \textit{effective information matrix}. Starting from phenomenological analysis, Tononi provided evidence for the theory by comparing the related model with neuroscientific results.

Psychological studies provided evidence in the general mechanisms of consciousness by employing experimental methodologies, such as \textit{response priming} and verbal reports. Unfortunately, experimental approaches which try to find the general functioning of the phenomenon failed in explaining the subjective effects. Explaining consciousness exclusively through general effects in the species can lead to the description of a \textit{philosophical zombie}, rather than of a human being. Evidence for this thesis was provided by Haggard \cite{haggard2008human}, who found that human beings often report experiences that do not correspond to their actual behavior or brain activity. 

In Medicine, the phenomenon is often associated with the concept of attention, as a measure of the subject's responsiveness, by giving to consciousness the general meaning of measuring cognitive abilities. In the present study, stimuli altering states of consciousness were not considered, since it was intended to investigate, from a computational perspective, the essential characteristics of the phenomenon.

To propose a model of computational consciousness, the concepts of objective/subjective and phenomenal/access consciousness seem fundamental, as they categorize the features of the phenomenon, even though at a high layer, with suitable consistency. Even though Neuroscience provides important insights regarding the neural activity of cognition, psychological studies focus on acquiring human experiences, describing consciousness with suitable abstraction from the brain. Therefore, the present study was intended to focus on the attempt of mathematical modeling some psychological mechanisms of the phenomenon, accepting the categories of objective/subjective and phenomenal/access consciousness, as well as the perspective proposed by Graziano and Kastner \cite{graziano2011human}, which conceive the process in terms of social perception. Concerning the assumption of consciousness as information integration, the present work considers Tononi's theory as consistent with the proposed model. In fact, as it can be evaluated in Section 4, the present work conceives consciousness as the capacity of a system to connect a given experience with all the other previously acquired experiences, by employing, to make a comparison with the Integration Information Theory, a sort of integration.

\section{Related work}
\label{sec:related_work}

One of the most relevant consciousness-inspired computational models of cognition is LIDA (Learning Intelligent Decision Agent) \cite{franklin2013lida, kugele2021learning}. The system is based on the Global Workspace Theory (GWT) \cite{baars1988cognitive}, which is a theory of consciousness considering a working memory of perceptual, evaluative, and attentional contents for activating computational conscious and unconscious processes. Unconscious cognition competes to access a \textit{Global Workspace} for entering consciousness; then, the relevant information is broadcast to motor systems to perform actions. At a given instant, the event the agent attends is processed in parallel by specialized modules; conscious information is globally available by the whole set of cognitive functions.

The GWT framework was recently adopted by Huang, Chella, and Cangelosi \cite{huang2022design}, which proposed a general model of computational consciousness based on top-down and bottom-up attention mechanisms (see Section \ref{sec:attention}) as criteria to access the Global Workspace. The authors assumed that not all the specialized modules of cognition are necessary for achieving consciousness, proposing a subdivision of the Global Workspace into separated nodes. By adopting convolutional neural networks for the visual and auditory stimuli acquisition and recurrent neural networks for action generation, they provided evidence for the machine to reproduce the \textit{attentional shift} \cite{sato2016neural} and \textit{blinking} \cite{raymond1992temporary} effects, together with the \textit{lag-1 sparing} \cite{hommel2005lag}. The experiments were conducted on simple visual and auditory stimuli, accounting for the above three fundamental phenomena of objective consciousness. The authors proposed their solution as a starting point for building further cognitive processes concerning subjective mechanisms of consciousness, such as emotional activity and short-term memory.

Lewis \cite{lewis2020rough} focused the modeling of computational consciousness on the concept of morality, proposing a mathematical tool based on the rough set theory and the Riemannian covariance matrix. It revealed possible to acquire data for extracting classifiers across many dimensions involving human moral behavior. As described by the author, an example of space involving significant dimensions is the plane in which the axes are \textit{moral act} and \textit{loving environment}, representing the covariance between positive and negative actions to environments. The relationship between positive and negative semantics of human behavior can be studied through rough set dimensions concerning the consciousness of a person or community.

The above solutions provided significant results for developing machine consciousness, but, to our knowledge, no suitable investigation has been conducted in AC concerning the functioning of consciousness as dependent on the relationship between emotional activity and morality. In the present study, emotions were considered fundamental for modeling consciousness; thus, important subjective mechanisms of machine consciousness can be obtained by introducing artificial emotional activity. As in the work conducted by Huang et al. \cite{huang2022design}, attention is assumed as a significant modulating factor for consciousness to happen. Similar to LIDA, the present study focuses on the integration of different parts of human cognition, but the correspondence between sensory stimuli acquisition and consciousness is assumed sequential, rather than parallel, i.e. cognitive tasks are executed sequentially.

\section{Background and assumptions}
\label{sec:prodromal_study}

The present study is based on a computational model of artificial cognition, mainly focused on the artificial reproduction of human emotional activity, which is called \textit{Smart Sensing} \cite{iovane2020smart}. The model was built by empirically assuming that human intelligence can be mathematically modeled as a hierarchy of cognitive layers concerning the concepts of sensation, perception, emotion, affection, attention, awareness, and consciousness. Figure \ref {fig:cognitive_model} shows a qualitative representation of the reference model in which the arrow symbol indicates a direct dependence between two cognitive layers. In the prodromal study \cite {iovane2020smart}, the authors specified a linguistic distinction between the concept of human cognitive layer and its relative artificial representation using mention with the initial capital letter (e.g., calling \textit {sensation} the human sensation, while \textit {Sensation} its modeled version); in the present study, it was decided to deploy the same linguistic artifice to keep the discussion clearer. 

Smart Sensing includes the modeling of the cognitive layers of Sensation, Perception, Emotion, and Affection, proposing the cognitive layers of Attention, Awareness, and Consciousness as future developments. The model acquires visual stimuli in the form of images through a convolutional neural network to produce, at each discrete time instant $n$, the cognitive instances $\textbf{r}_{1,n}$, $\textbf{r}_{2,n}$, $\textbf{r}_{3,n}$, and $\textbf{r}_{4,n}$ associated with the cognitive layers of Sensation, Perception, Emotion, and Affection, respectively. 

The Sensation cognitive instance $\textbf{r}_{1,n} \in \mathbb{R}^{1 \times k}$ is defined as

\begin{equation}
    \textbf{r}_{1,n} = D_{1}(n, \textbf{m}_{1,n},\textbf{x}_{n}, t) = \textbf{k}_{1,n} \bigg(\alpha_{1} thr_{H}(\textbf{x}_{n}) \frac{e^{-\frac{t-n}{n}} - e^{-\frac{T_{1}}{n}}}{{1- e^{-\frac{T_{1}}{n}}}} + \beta_{1} \textbf{m}_{1,n} \bigg),
\end{equation}

\begin{equation}
t_{b}<n \implies D_{1}(n, \textbf{m}_{1,n},\textbf{x}_{n}, t_{b}) = 0 \qquad n \in \mathbb{N},
\end{equation}

\begin{equation}
t_{b}>n+T_{1} \implies D_{1}(n, \textbf{m}_{1,n},\textbf{x}_{n}, t_{b}) = 0 \qquad n \in \mathbb{N},
\end{equation}

\begin{equation}
    \textbf{m}_{1,n} = \begin{pmatrix}
    D_{1}(n_{f}, \textbf{m}_{1,n_{f}}, \textbf{x}_{n_{f}}, n)\\
    \vdots\\
    D_{1}(n_{g}, \textbf{m}_{1,n_{g}}, \textbf{x}_{n_{g}}, n)\\
    \vdots\\
    D_{1}(n_{h}, \textbf{m}_{1,n_{h}}, \textbf{x}_{n_{h}}, n)\\
  \end{pmatrix} \qquad n_{f}<n_{g}<n_{h}<n \in \mathbb{N},
\end{equation}

where $\textbf{x}_{n} \in \mathbb{R}^{1 \times k}$ is the input at the instant $n$, while $T_{1} > 0$, $\alpha_{1} \in \mathbb{R}^{1 \times k}$, and $\beta_{1} \in \mathbb{R}^{1 \times l}$ are the removal period, the weights of the current input, and the weights of the memory $\textbf{m}_{1,n} \in \mathbb{R}^{l \times 1}$ of the Sensation cognitive layer, respectively, with $k$ the input dimensions and $l = n_{h} - n_{f} +1$ the number of elements in the memory. The function $D_{1}$ represents the trend through which the cognitive instances decay exponentially with the time $t$. The $thr_{H}(\cdot)$ function sets to zero all the components of the input $\textbf{x}_{n}$ that are lower than the threshold $H$. Finally, it holds that $\textbf{k}_{1,n} = \textbf{r}_{2,n-1}\textbf{r}_{5,n-1}$, where $\textbf{r}_{2,n-1} \in \mathbb{R}^{1 \times k}$ and $\textbf{r}_{5,n-1} \in \mathbb{R}^{1 \times k}$ represent the Perception and Attention cognitive instances at the instant $n-1$, respectively.

The Perception, Emotion, and Affection cognitive instances, i.e., the instances $\textbf{r}_{i,n} \in \mathbb{R}^{1 \times k}$, with $i \in \{2,3,4\}$, are defined as

\begin{equation}
    \textbf{r}_{i,n} = D_{i}(n, \textbf{m}_{i,n},\textbf{x}_{n}, t) = \textbf{k}_{i,n} \textbf{A}_{i}\bigg(\frac{{\alpha_{i}}\textbf{x}_{n}}{t} + \frac{\beta_{i}\textbf{m}_{i,n}}{t^2} - t_{0_{i}}\bigg),
\end{equation}

\begin{equation}
    \textbf{A}_{i} = \frac{{\alpha_{i}}\textbf{x}_{n} + {\beta_{i}} \textbf{m}_{i,n}}{\frac{\alpha_{i}\textbf{x}_{n}}{t} + \frac{\beta_{i} \textbf{m}_{i,n}}{t^2} - t_{0_{i}}},
\end{equation}

\begin{equation}
    t_{0_{i}} = \frac{{\alpha_{i}}\textbf{x}_{n}}{t+T_{i}} + \frac{\beta_{i} \textbf{m}_{i,n}}{(t+T_{i})^2},
\end{equation}

\begin{equation}
    \textbf{m}_{i,n} = \begin{pmatrix}
    D_{i}(n_{f}, \textbf{m}_{i,n_{f}}, \textbf{x}_{n_{f}}, n)\\
    \vdots\\
    D_{i}(n_{g}, \textbf{m}_{i,n_{g}}, \textbf{x}_{n_{g}}, n)\\
    \vdots\\
    D_{i}(n_{h}, \textbf{m}_{i,n_{h}}, \textbf{x}_{n_{h}}, n)\\
  \end{pmatrix} \qquad n_{f}<n_{g}<n_{h}<n \in \mathbb{N},
\end{equation}

where $\textbf{x}_{n}$ is the input at the instant $n$, while $T_{i} >0$, $\alpha_{i} \in \mathbb{R}^{1 \times k}$, and $\beta_{i} \in \mathbb{R}^{1 \times l}$, are the removal period, the weights of the current input, and the weights of the memory $\textbf{m}_{i,n} \in \mathbb{R}^{l \times 1}$ of the $i$-th cognitive layer, respectively, with $i \in \{2,3,4\}$, $k$ the input dimensions, and $l = n_{h} - n_{f} +1$ the number of elements in the $i$-th memory. The function $D_{i}$ represents the trend through which the cognitive instances decay polynomially with the time instance $n$. Finally, it holds that $\textbf{k}_{2,n} = \textbf{k}_{4,n} = \textbf{j}$, with $\textbf{j} \in \mathbb{R}^{1 \times k}$ representing a unitary vector, while $\textbf{k}_{3,n} = \textbf{r}_{4,n-1}$, where $\textbf{r}_{4,n-1} \in \mathbb{R}^{1 \times k}$ is the Affection cognitive instance at the instant $n-1$.

Each $i$-th cognitive layer, with $i \in \{1,2,3,4\}$, computes the instances by considering its own memory $\textbf{m}_{i,n}$, called the \textit{cognitive memory}. Sensation's cognitive memory $\textbf{m}_{1,n}$ is characterized by an exponential decay function, while the memories $\textbf{m}_{2,n}$, $\textbf{m}_{3,n}$, and $\textbf{m}_{4,n}$ associated with the cognitive layers of Perception, Emotion, and Affection are described polynomially. These memories consist in time windows storing decaying cognitive instances to reproduce the concept of short-term memory artificially. Each cognitive instance decreases in its relevance in the function of the time $n$ to a predefined removal period $T_{i}$, i.e., with respect to the time interval in which a cognitive instance is stored in the relative memory. 

To associate the cognitive instances provided by the Emotion with the classes of happiness, anger, fear, surprise, contempt, sadness, disgust, and the neutral state, the authors trained a learner on three different episodes depicting visual scenes eliciting happiness (i.e., “beautiful woman” and “own home”), anger (i.e., “murder of animal”), fear (i.e., “war,” “man pointing weapon,” and “terrorism”), surprise (i.e., “crazy sportsman”), contempt (i.e., “politician” and “parking car”), sadness (i.e., “someone's death or sick” and “car accident”), disgust (i.e., “injury” and “autopsy”), and the neutral state (i.e., “landscape”).

For the training of the learner, the authors associated the Emotion cognitive instance at the discrete time $n$ with a pre-defined emotional class in the dataset. For instance, in the case the stimuli at the time instants $n-2$, $n-1$, and $n$ are “murder of animal” (labeled with anger), “war” (labeled with fear), and “crazy sportsman” (labeled with surprise), respectively, the emotional class associated with the instant $n$ will be the surprise. However, the above orientation of the model is not mandatory and it depends on the emotional history it was intended to provide to the agent. The classification of Emotion cognitive instances introduces an element of emotional subjectivity in the model.

\begin{figure}[!h]
    \centering
    \includegraphics[width=0.7\textwidth]{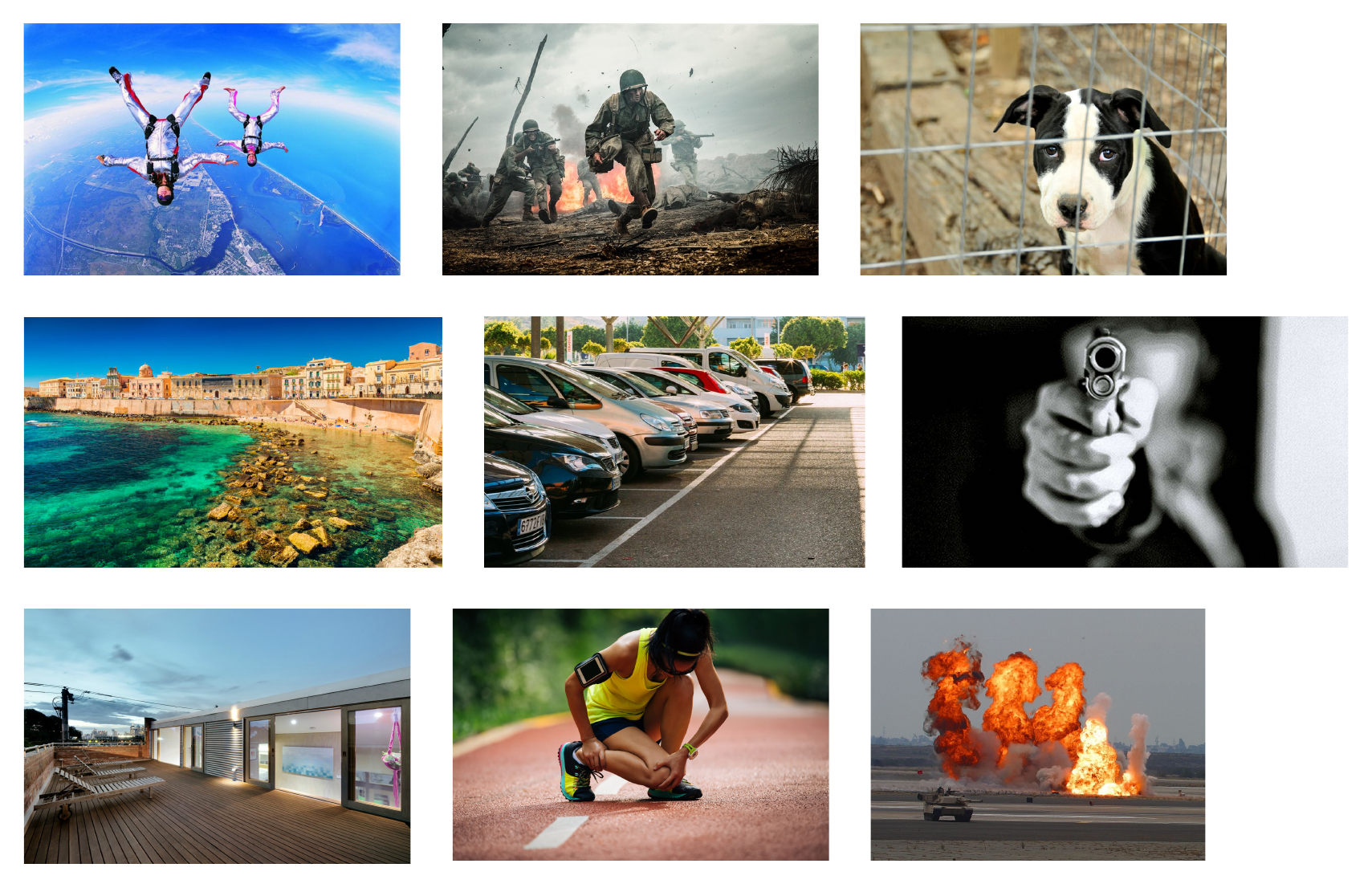}
    \caption{Some of the images employed in the custom dataset for training the learner.}\label{fig:samples_images}
\end{figure}

During experimentation, by adopting VGG16 \cite{simonyan2014very} pre-trained on ImageNet \cite{deng2009imagenet} for feature extraction and XGBoost \cite{chen2016xgboost} as the learner for associating Emotion cognitive instances to the emotional classes and the neutral state, the authors achieved 85\% accuracy on a custom dataset composed of 612 images, of which the 80\% was employed for the training phase. To provide some instances of the above dataset, Figure \ref{fig:samples_images} shows some of the images employed for training the learner.

Subsequently, by evaluating the model on different episodes (the process of \textit{emotional activity}, as called by the authors), they found a lowering in the accuracy score as the $\alpha_{3}$ weights of the Emotion cognitive instance increased in magnitude \cite{iovane2020smart}. The same evaluation provided an increase in the accuracy score as the $\alpha_{3}$ weights decreased towards zero. The above test proved that Smart Sensing reproduces certain phenomena of human emotional activity, artificially. Similar to a human being, the agent evaluates an emotional history, i.e., a series of stimulus-emotion associations which leverage emotional activity concerning different episodes of stimuli. The model represents an extreme reduction of human cognition, but it allows to reach, as described in the prodromal study, a coherent approximation of the evidence found in the area of Cognitive Psychology and Neuroscience. In their work, the authors compared the functioning of the model with the related experimental evidence involving human beings. 

The present study continues the development of the above proposed model of computational consciousness by re-evaluating the Smart Sensing and modeling the cognitive layers of Attention, Awareness, and Consciousness.

\section{Proposed model of computational consciousness}
\label{sec:proposed_model}

\begin{figure}[!h]
    \centering
    \includegraphics[width=0.4\textwidth]{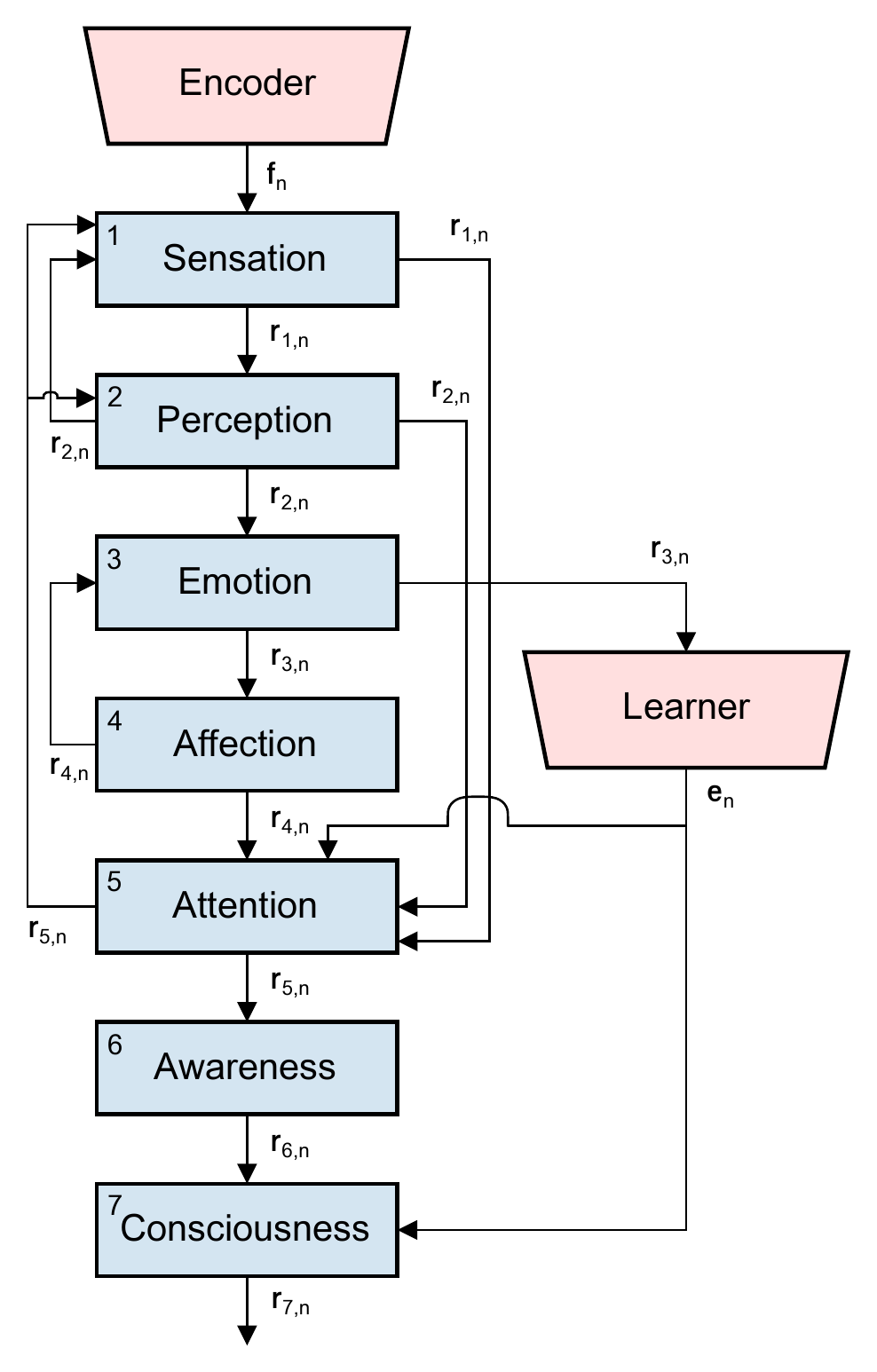}
    \caption{A qualitative representation of the proposed model of computational consciousness.}\label{fig:cognitive_model}
\end{figure}

The proposed model acquires sensory stimuli by sampling occurrences from the environment as a consequence of evaluating the visual artificial channel. As shown in Figure \ref{fig:cognitive_model}, the feature vector $\textbf{f}_{n} \in \mathbb{R}^{1 \times k}$, with $k$ the input dimensions, is processed through a hierarchy of seven cognitive layers, i.e., Sensation, Perception, Emotion, Affection, Attention, Awareness, and Consciousness, which compute cognitive instances and pass temporary results to other cognitive layers. 

The dependencies concerning cognitive layers can be listed as follows.

\begin{itemize}
\item  Sensation depends on Attention and Perception: as the cognitive instances of Attention increase in magnitude, the model acquires sensory occurrences more intensely; as the cognitive instances of Perception increase in magnitude, the model provides higher significance to the acquired stimuli.
\item Perception depends on Sensation: to produce cognitive instances, Perception acquires temporary results from the Sensation cognitive layer.
\item Emotion depends on Perception and Affection: to produce cognitive instances, Emotion acquires temporary results from Perception and provides higher significance to the produced cognitive instances as Affection increases in magnitude.
\item Affection depends on Emotion: to produce cognitive instances, Affection acquires temporary results from the Emotion cognitive layer.
\item Attention depends on Sensation, Perception, Emotion, and Affection: to produce cognitive instances, Attention acquires temporary results from the Sensation, Perception, and Affection cognitive layers, while the probability vector $\textbf{e}_{n} \in [0,1]^{1 \times C_{e}}$ related to the classification of the current Emotion cognitive instance from a learner, with $C_{e}$ the quantity of considered emotional classes.
\item Awareness depends on Attention and observes Sensation, Perception, Emotion, and Affection cognitive instances: to produce cognitive instances, Awareness acquires temporary results from Attention and state representations of Sensation, Perception, Emotion, and Affection cognitive layers.
\item Consciousness depends on Awareness and connects Sensation, Perception, Emotion, and Affection cognitive instances: to produce cognitive instances, Consciousness acquires temporary results from Awareness and connects state representations of Sensation, Perception, Emotion, and Affection through semantics produced from classifications of Emotion cognitive instances provided by a learner.
\end{itemize}

As explained in Section \ref{sec:attention}, the Attention cognitive layer was assumed as dependent directly on Sensation, Perception, and Affection, since it processes cognitive instances, while it depends on Emotion through the classifications of cognitive instances provided by the deployed learner. 

Awareness was assumed as dependent directly on the Attention to represent the feature suggested by Huang et al. \cite{huang2022design}, for which the process of awareness depends on attention, while not dependent directly on Sensation, Perception, Emotion, and Affection. Awareness does not process cognitive instances but states representations concerning the whole set of cognitive layers preceding Attention. As it can be verified in Section \ref{sec:awareness}, the representation above is the information fusion of probability, plausibility, credibility, and possibility scores associated with the current processing stimuli. This feature was introduced to highlight the value of the intuition provided by Graziano and Kastner \cite{graziano2011human}, which assume the awareness as the perception of an entity which is, in turn, aware of stimuli acquired from the environment and body. To make a comparison with our proposal of a computational model of consciousness, the Awareness cognitive layer processes a stimulus by objectifying the Sensation, Perception, Emotion, and Affection cognitive instances, i.e., by perceiving them as external, through the evaluation of state representations. 

Finally, Consciousness was assumed as dependent directly on Awareness but indirectly on Sensation, Perception, Emotion, and Affection. This cognitive layer connects the state representations provided by the Awareness and associates them with moral semantics which depend on the classifications of Emotion cognitive instances provided by the deployed learner. As it can be verified in Section \ref{sec:consciousness}, the connection of states representations provided by the Awareness cognitive layer, together with the related semantics associations, was modeled through the adjacency matrix of a graph.

Under the above assumptions, the following Sections describe the part of the model concerning cognitive layers of Attention, Awareness, and Consciousness.

\subsection{Attention}
\label{sec:attention}

Through sensations, the human being confronts reality with the continuous perception of stimuli, which are subjected to selection based on their importance. In particular, when listening to an interesting lesson, the visual and auditory systems focus on capturing the information that comes from the outside, making the understanding of the proposed contents easier. The above process is called \textit{attention}, which can be defined as the cognitive process of selecting the stimuli received through the five senses, or as the ability to focus on specific information coming from the environment or the body. Attention is often intended as the efficient management of cognitive resources, as the brain would be characterized by a limited capacity to process all the information at a given time. This definition is not far from the concept of \textit{inattentional blindness} related to visual attention, which considers the physical inability to process all the information coming from the environment. In fact, at any time, the human being misses a substantial part of the visual world \cite{mack2003inattentional}.

Visual attention can be declined into \textit{covert} and \textit{overt}. The first refers to the act of processing a given stimulus without moving the eyes, while the second concerns the action of selectively focusing on a particular position of interest \cite{findlay2003active}. In overt attention, the distinction between reflexive and controlled eye movements is significant: reflexive attention is characterized by the involuntary focus following a stimulus of particular relevance, while controlled attention employs the voluntary selection of a visual stimulus. More generally, an attention process affected exclusively by external factors is often called \textit {bottom-up attention}, while that determined by internal factors, intended as prior acquired knowledge, is often called \textit {top-down attention} \cite{katsuki2014bottom}. Similar to the definition of overt attention is the \textit{selective attention}, which consists of focusing on a part, while neglecting the rest, of a stimulus of interest. This definition is also used in association with the sensory sources different than the visual channel, such as the auditory perception \cite{ouchi2015normal}. The attention involves all the senses, including touch \cite{spence2007recent}, taste, and smell \cite{veldhuizen2011modality}; the overall attention of a subject can be considered as a combination of focuses on different sensory channels.

The above cognitive process was proved to be dependent on emotions, as it depends on the affective reaction elicited by a stimulus \cite{van2013disgust, van2014disgust}. Attention was found also dependent on the culture, as proved by Chavajay and Rogoff \cite{chavajay1999cultural}, but the present study wanted to focus on the general characteristics only.

Focusing on the visual sensation, Attention can be considered as a mind cognitive process that allows to select or ignore environmental stimuli. Therefore, the related cognitive instance $\textbf{r}_{5,n} \in [0,1]^{1 \times 1}$ is dependent on time and on the cognitive matrix $\textbf{A}_{c_{n}} \in [0,1]^{4 \times k}$. Formally,

\begin{equation}
    \textbf{r}_{5,n} = ten(\textbf{A}_{c_{n}}),
\end{equation}

with

\begin{equation}
           \textbf{A}_{c_{n}} = \begin{pmatrix}
    \mathring{\textbf{s}}_{n}\\
    \mathring{\textbf{p}}_{n}\\
    \textbf{E}_{s} \\
    \mathring{\textbf{a}}_{n}
    \end{pmatrix} =  
    \begin{pmatrix}
    a_{11_{n}} & \hdots & a_{1k_{n}}\\
    a_{21_{n}} & \hdots & a_{2k_{n}}\\
    a_{31_{n}} & \hdots & a_{3k_{n}}\\
    a_{41_{n}} & \hdots & a_{4k_{n}}
    \end{pmatrix},
\end{equation}

\begin{equation}
    \mathring{\textbf{s}}_{n} = \frac{\textbf{r}_{1,n}}{S_{b}},\qquad \mathring{\textbf{p}}_{n} = \frac{\textbf{r}_{2,n}}{P_{b}},\qquad
\mathring{\textbf{a}}_{n} = \frac{\textbf{r}_{4,n}}{A_{b}},
\end{equation}

\begin{equation}
    ten(\textbf{A}_{c_{n}}) = \frac{\max\limits_{j \in \{1,...,k\}}\{\max\limits_{i \in \{1,...,4\}}\{a_{ij_{n}}\}\} + \underset{H_{c} = 1}{\sum_{i=1}^{4} \sum_{j=1}^{k}}a_{ij_{n}}}{2},
\end{equation}

where $ten(\cdot)$ is called the \textit{tension} function. The vector $\textbf{E}_{s} \in [0,1]^{1 \times k}$ contains the probability vector $\textbf{e}_{n}$ provided by the learner, i.e., the relative probabilities associated with the classification of the Emotion cognitive instance at the discrete time $n$, zero-padded with $k-C_{e}$ elements. Vectors $\mathring{\textbf{s}}_{n}$, $\mathring{\textbf{p}}_{n}$, and $\mathring{\textbf{a}}_{n}$ are the normalized versions of the Sensation, Perception, and Affection cognitive instances. Quantities $S_{b}$, $P_{b}$, and $A_{b}$ are the upper bounds that the components of the Sensation, Perception, and Affection cognitive instances can reach, respectively. Finally, the $H_{c}$ parameter represents a saturation threshold; since $\textbf{A}_{c_{n}}$ is a matrix defined in $[0,1]^{4 \times k}$, the bound $H_{c} = 1$ was chosen.

The cognitive layer of Attention processes the normalized distributions of cognitive instances to provide the related tension. For instance, in the case the agent receives a sensory impulse, i.e., when the cognitive matrix provides, except for a single component close to $1$, all equally distributed, and close to zero, occurrences, the magnitude of the Attention cognitive instance increases accordingly. The above model can be compared with the category of bottom-up attention.

\subsection{Awareness}
\label{sec:awareness}

Awareness represents that layer of cognition through which a human being acquires experience of him/herself and the surrounding environment. Many use the term “awareness” as a synonym for consciousness, intending awareness as “becoming conscious” of something. However, the present study considers the above assumption is not necessarily true, as consciousness would represent the process that associates awareness to moral semantics. One can be aware of an event, such as the breaking of a glass or the sound of the breath, without attributing meanings to these stimuli. In the present study, it was assumed that acquired stimuli become associated with semantics only through consciousness. The phenomenon of awareness can be conceived as the acquisition of experiences that come from the outside, such as the perception of a sensation, or the inside, as an emotion \cite{hussain2009brain}. Awareness seems to be a sort of cognitive relevance of a human's sensory, perceptive, emotional, and affective experience. Consciousness, on the other hand, would consist in the association of the aforementioned experiences with positive, negative, or neutral moral semantics.

Awareness can be conceived as the cognitive reaction to the occurrence of certain conditions or given events. It was assumed as a function of time and the Attention cognitive layer. In the case the stimuli are exclusively internal, the awareness is regarded as “self-awareness” or “subjective awareness” since, for instance, a subject can be aware of something under subconscious perspective, i.e., gaining awareness through internal states, visceral sensations, or sensory perceptions related to the external events. The above faculty seems to be related to the comprehension of environmental and subjective events in terms of intensity. The \textit{Environment Awareness} can be conceived under the statement “I know,” while the \textit{Subjective Awareness} with the statement “I am.” Consciousness, instead, seems to be a “deeper” layer of awareness enriched by the semantics of social and personal morality. The \textit{Environment Consciousness} can be synthesized with the statement “I evaluate,” while the \textit{Subjective Consciousness} with the statement “I am me.” Awareness was modeled as the cognitive layer that provides the agent with objective knowledge about external and subjective events. Consciousness was assumed to be activated after the processing of the above pool of knowledge to enrich the Awareness cognitive instances with critical evaluations. Under the above hypotheses, it was possible to base the modeling of Awareness and Consciousness by adopting the following scheme:

$$\underset{\textit{Environment Awareness}}{\textit{I know}} \implies \underset{\textit{Environment Consciousness}}{\textit{I evaluate}}$$
$$\underset{\textit{Subjective Awareness}}{\textit{I am}} \implies \underset{\textit{Subjective Consciousness}}{\textit{I am me}}.$$

In the proposed model, Environment Awareness was associated with the knowledge, analysis, and inference concerning the occurrence of external events; more specifically, Sensation and Perception cognitive instances represent the knowledge acquired by the agent. Subjective Awareness, instead, was associated with the knowledge related to subjective events; in this case, Emotion and Affection cognitive instances represent the knowledge acquired by the agent.

The theory of decision and reasoning in info-incompleteness conditions \cite{iovane2020decision} was adopted for modeling the influence of others (humans or agents) on personal opinions or experiences. It was possible to describe the Awareness in terms of probability, plausibility, credibility, and possibility associated with Sensation, Perception, Emotion, and Affection cognitive instances. This approach introduces into the model the awareness that the agent can reach as a consequence of an external suggestion, i.e., awareness that does not depend on direct experience (plausibility, credibility, and possibility scores, in the model). For instance, the artificial agent can acquire awareness either by directly processing the occurrences of a given event or by extracting information from the experiences of other agents or human beings. Additional Sensation, Perception, Emotion, and Affection cognitive instances, together with the relative plausibility, credibility, and possibility scores can be acquired by the agent as a consequence of the interaction with the environment.

Hence, at the instant $n$, for each cognitive instance $\textbf{r}_{i,n}$, with $i \in \{1,2,3,4\}$, the agent computes the distributions

\begin{equation}
    Pr(\textbf{r}_{i,n}): \mathbb{R}^{1 \times k} \rightarrow [0,1],\end{equation}

\begin{equation}
    Pl(\textbf{r}_{i,n}): \mathbb{R}^{1 \times k} \rightarrow [0,1],
\end{equation}

\begin{equation}
    Cr(\textbf{r}_{i,n}): \mathbb{R}^{1 \times k} \rightarrow [0,1],
\end{equation}

\begin{equation}
    Po(\textbf{r}_{i,n}): \mathbb{R}^{1 \times k} \rightarrow [0,1],
\end{equation}

where $Pr(\textbf{r}_{i,n})$, $Pl(\textbf{r}_{i,n})$, $Cr(\textbf{r}_{i,n})$ and $Po(\textbf{r}_{i,n})$ are, respectively, the probability, plausibility, credibility, and possibility related to a given cognitive instance $\textbf{r}_{i,n}$ at the discrete instant $n$.

The information fusion of the above distributions provides the expectation function 

\begin{equation}
  O_{i}(n)=\begin{cases}
    Pr(\textbf{r}_{i,n}) \quad \text{if $0.05 \leq Pr(\textbf{r}_{i,n}) \leq 1.00$,}\\
    (Pr(\textbf{r}_{i,n}) + w_{1,i}Pl(\textbf{r}_{i,n}))/2 \quad \text{if $0.01 < Pr(\textbf{r}_{i,n}) \leq 0.05$,}\\
    (Pr(\textbf{r}_{i,n}) + w_{1,i}Pl(\textbf{r}_{i,n}) + w_{2,i}Cr(\textbf{r}_{i,n}))/3 \quad \text{if $0.005 < Pr(\textbf{r}_{i,n}) \leq 0.01$,}\\
    (Pr(\textbf{r}_{i,n}) + w_{1,i}Pl(\textbf{r}_{i,n}) + w_{2,i}Cr(\textbf{r}_{i,n}) + w_{3,i}Po(\textbf{r}_{i,n}))/4 \quad \text{if $Pr(\textbf{r}_{i,n}) < 0.005$},
  \end{cases}
\end{equation}

where $w_{j,i} \in [0,1]$, with $j=\{1,2,3\}$, are the weights related, respectively, to the plausibility, credibility, and possibility associated with the $i$-th cognitive layer.

Cognitive instances probabilities are computed through a frequentist approach, i.e., by counting the occurrences of a given instance in the function of time. Since similar sensory stimuli can result in non-stackable cognitive instances, it was useful to support the above process by means of similarity measures. At each step, for each cognitive layer of the Smart Sensing, the Awareness computes the Euclidean distance between the current acquired cognitive instance and all the previously captured occurrences. Then, the counter associated with the current cognitive instance is incremented only in the case the above distance results greater than or equal to a predefined threshold $R$. Formally,

\begin{equation}
    \begin{cases} C(\textbf{r}_{i,n}) = C(\textbf{r}_{i,m})+1 \leftrightarrow \exists \; \textbf{r}_{i,m} \in M_{i, n-1}: \; 0 \leq d(\textbf{r}_{i,n}, \textbf{r}_{i,m}) \leq R\textrm{,} \\
C(\textbf{r}_{i,n}) = 1 \quad \textrm{otherwise,}
\end{cases}
\end{equation}

where

\begin{equation}
    Pr(\textbf{r}_{i,n}) = \frac{C(\textbf{r}_{i,n})}{C_{n}},
\end{equation}

with $m < n\in \mathbb{N}$, $R > 0$ the reference distance threshold, $C_{n}$ the count of the cognitive instances at the instant $n$, and $M_{i,n-1}$ the set of the reference cognitive instances acquired up to the instant $n-1$, i.e., all the instances which counters are greater than or equal to 1. The terms $C(\textbf{r}_{i,n})$ and $C(\textbf{r}_{i,m})$ are the counters related to the current and reference cognitive instances, respectively. Finally, the term $d(\textbf{r}_{i,n}, \textbf{r}_{i,m})$ represents the Euclidean distance between the above instances.

Theoretically, cognitive instances plausibility was defined to be determined by experts' opinions, credibility by affectively relevant identities opinions, and possibility by sentiment analysis results. With this approach, the agent can enrich its cognition even when it does not directly experience sensory stimuli or when the number of occurrences is not suitably high. 

Under the above hypotheses, the Awareness cognitive instance $\textbf{r}_{6,n} \in \mathbb{R}^{1 \times 1}$ was defined as the sum of the Environmental and Subjective Awareness. Formally,

\begin{equation}
    \textbf{r}_{6,n} = \textbf{k}_{6,n}(\underbrace{\phi_{1}O_{1}(n) + \phi_{2}O_{2}(n)}_{\textit{Environment Awareness}} + \underbrace{\phi_{3}O_{3}(n) + \phi_{4}O_{4}(n)}_{\textit{Subjective Awareness}}),
\end{equation}

where $O_{1}(n)$, $O_{2}(n)$, $O_{3}(n)$, and $O_{4}(n)$ are the Sensation, Perception, Emotion, and Affection expectation functions and $\phi_{1}$, $\phi_{2}$, $\phi_{3}$, and $\phi_{4}$ their related weights, all defined in $\mathbb{R}^{1 \times 1}$, respectively.

\begin{figure}[!h]
    \centering
    \includegraphics[width=0.9\textwidth]{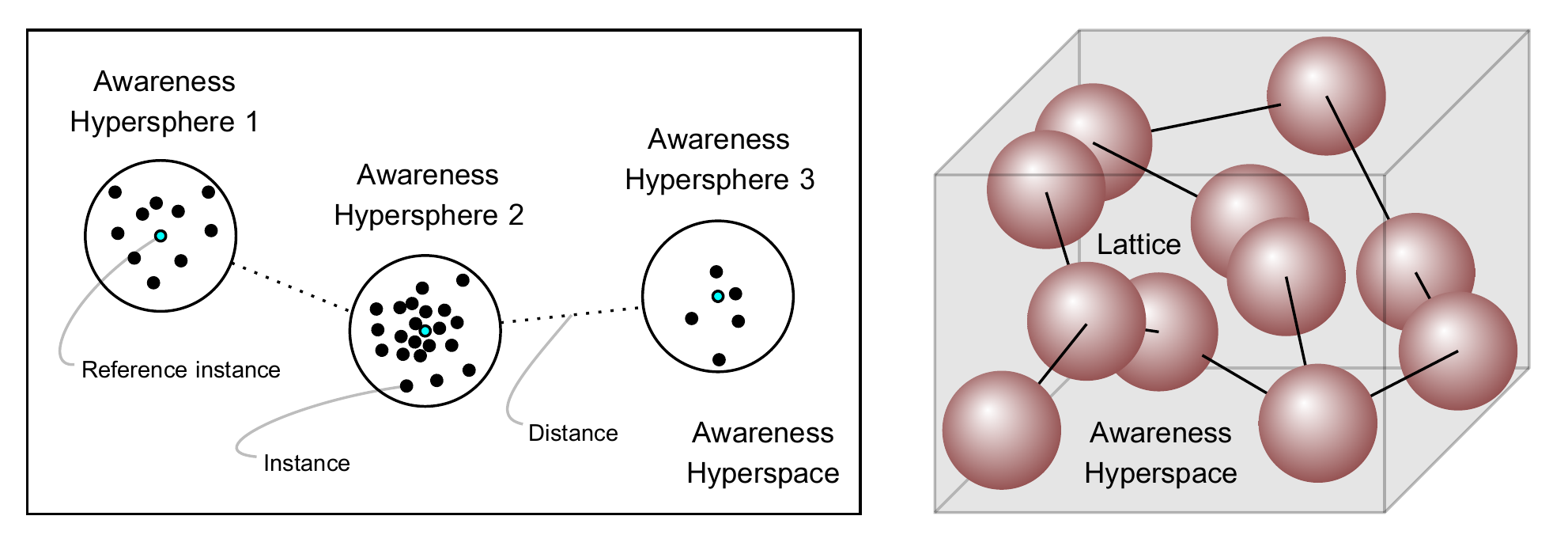}
    \caption{On the left, a qualitative 2D representation of the Awareness Hyperspace; the centers of the hyperspheres represent the reference cognitive instances $\textbf{r}_{i,m} \in M_{i, n-1}$ and dotted lines their relative distances. On the right, a qualitative 3D representation in the form of a lattice.}\label{fig:hyperspheres}
\end{figure}

Awareness is supported by a geometrical model, in which the distribution of cognitive instances, as shown in Figure \ref{fig:hyperspheres}, is encapsulated into hyperspheres, with ray $R$, in which the centres are the reference instances $\textbf{r}_{i,m} \in M_{i,n-1}$. Formally,

\begin{equation}
    \hat{V}_{\textbf{r}_{i,m}} = \{\textbf{r}_{i,n} \in AH_{i} : \; 0 \leq d(\textbf{r}_{i,n}, \textbf{r}_{i,m}) \leq R\},
\end{equation}

where $AH_{i} \in \mathbb{R}^{k}$ is the Awareness Hyperspace associated with the $i$-th cognitive layer, with $k$ the dimensions of cognitive instances. 

By assuming the hyperspheres as non-intersecting, it is possible to conceive the Awareness as similar to an amorphous crystalline lattice in which the molecules are the hyperspheres themselves. In the proposed model, the intensity of Awareness related to a given reference cognitive instance is considered at its maximum when the hypersphere becomes completely, i.e., infinitely, full of instances. 

\subsection{Consciousness}
\label{sec:consciousness}

In the present study, the phenomenon of consciousness was considered as the semantic processing of awareness. Awareness was conceived as a cognitive process in which the experiences are acquired and distinguished, but not connected and not associated with semantics; Consciousness links the acquired experiences and associates them with semantics. Consciousness was not assumed as a process exclusively related to the acquisition of a stimulus, but conceived also as a process connecting a stimulus with the other previously acquired stimuli. For simplicity, the semantics of experiences, which in the present model were modeled as Awareness cognitive instances, can be interpreted as positive, negative, and neutral. The semantics related to the environment are hypothesized as affected by the social morality, i.e., the semantics that an agent learns from the external context; the semantics related to subjectivity are assumed as affected by the personal morality, which in the present study are obtained from the emotional activity (e.g., when an experience elicits a negative emotion, like fear, the relative subjective semantic associated with that experience is negative). 

Consciousness was modeled as the sum of contributions concerning the entropy and energy of an undirected graph, called the \textit{Consciousness Graph}, in which the nodes are associated with semantics. Two types of graphs were conceived: the first considers the Environment Awareness by deploying the hyperspheres in the Sensation and Perception Hyperspaces as nodes; the second considers the Subjective Awareness by deploying the hyperspheres in the Emotion and Affection Hyperspaces as nodes. Each hypersphere is represented through the relative reference instance, which union with a semantic provides a \textit{social morality instance} for the Environment Awareness, while a \textit{personal morality instance} for the Subjective Awareness. Entropy and energy represent two metrics of the complexity of information present in the graphs, in terms of connections among reference cognitive instances and positivity, negativity, and neutrality of the relative semantics. This approach permitted to model the consciousness of an experience as a relationship with other experiences. Furthermore, the data structure of the graph permits also to find paths from a reference cognitive instance to all the other reference cognitive instances.

\begin{figure}[!h]
    \centering
    \includegraphics[width=0.75\textwidth]{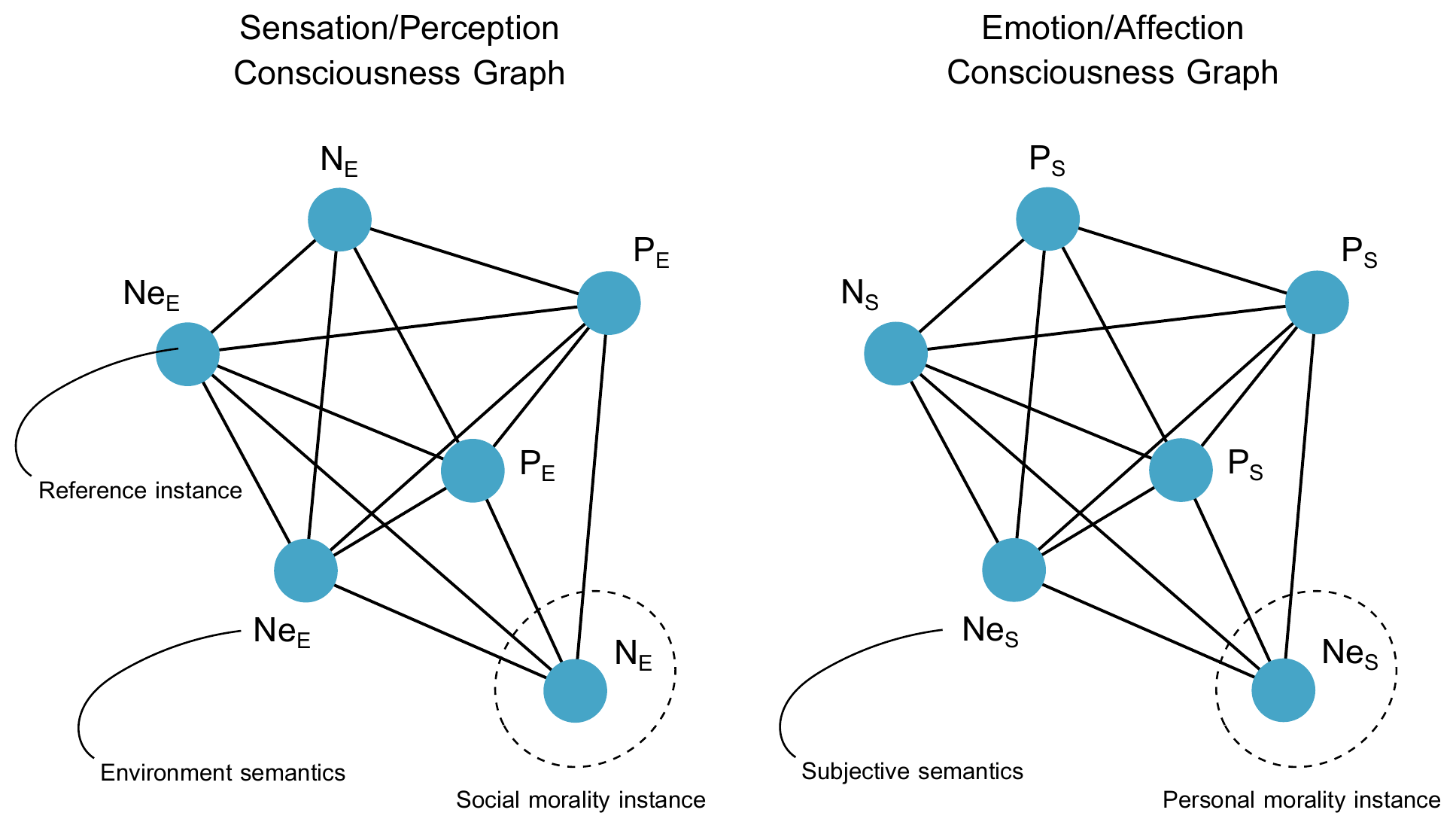}
    \caption{A qualitative representation of the Sensation, Perception, Emotion, and Affection Consciousness Graphs. $N_{E}$, $P_{E}$, and $Ne_{E}$ represent the negative, positive, and neutral environment semantics, while $N_{S}$, $P_{S}$, and $Ne_{S}$ the negative, positive, and neutral subjective semantics.}\label{fig:consciousness_graphs}
\end{figure}

Figure \ref{fig:consciousness_graphs} shows a qualitative example of the above two types of graphs. As it can be seen, the environment semantics, which are associated with the Sensation and Perception Consciousness Graphs can be different from the subjective semantics, which are associated with the Emotion and Affection Consciousness Graphs.

The Consciousness Graph $G_{i}$ is associated with an adjacency matrix $\textbf{J}_{i,n} \in \mathbb{R}^{m \times m}$, with $m$ the quantity of reference cognitive instances, i.e., the centers of hyperspheres, of generic element

\begin{equation}
g_{l,h} = \frac{S_{l} + S_{h}}{2} d(\textbf{r}_{i,l},\textbf{r}_{i,h}),
\end{equation}

where $l$ and $h$ are the indexes of the reference instances $\textbf{r}_{i,l}, \textbf{r}_{i,h} \in M_{i, n-1}$ concerning two adjacent nodes, with $S_{l}, S_{h} \in \{-1,0,1\}$ the related semantics. The term $d(\textbf{r}_{i,l}, \textbf{r}_{i,h})$ in the equation represents the Euclidean distance between two reference instances.

The Consciousness cognitive instance $\textbf{r}_{7,n} \in \mathbb{R}^{1 \times 1}$ depends directly on the Awareness and was defined as the sum of the Environmental and Subjective Consciousness. Formally,

\begin{equation}
    \textbf{r}_{7,n} = \textbf{k}_{7,n}(\underbrace{\psi_{1}Y_{1}(n) + \psi_{2}Y_{2}(n)}_{\textit{Environment Consciousness}} + \underbrace{\psi_{3}Y_{3}(n) + \psi_{4}Y_{4}(n)}_{\textit{Subjective Consciousness}}),
\end{equation}

with 

\begin{equation}
    Y_{i}(n) = d(A_{i}(0), A_{i}(n)),
\end{equation}

\begin{equation}
    A_{i}(n) = (H(J_{i,n}),E(J_{i,n})),
\end{equation}

\begin{equation}
    A_{i}(0) = (H(J_{i,0}),E(J_{i,0})),
\end{equation}

where $Y_{1}(n)$, $Y_{2}(n)$, $Y_{3}(n)$, and $Y_{4}(n)$ are the Sensation, Perception, Emotion, and Affection Consciousness intensities and $\psi_{1}$, $\psi_{2}$, $\psi_{3}$, and $\psi_{4}$ their related weights, all defined in $\mathbb{R}^{1 \times 1}$, respectively. The Consciousness intensity $Y_{i}(n)$ of the $i$-th cognitive layer is computed through the Euclidean distance between the point $A_{i}(n)$ and the origin of the entropy-energy plane associated with the adjacency matrix $\textbf{J}_{i,n}$.

The Consciousness energy of the $i$-th cognitive layer was defined as

\begin{equation}
    E(\textbf{J}_{i,n}) = \sum_{l,h} |\bar{g}_{l,h}|,
\end{equation}

where $\bar{g}_{l,h}$ is the generic element of the normalized version of the adjacency matrix $\textbf{J}_{i,n}$. This quantity increases proportionally to the distance between two reference instances and the positivity of their semantics in the related hyperspace.

The Consciousness entropy of the $i$-th cognitive layer was defined as

\begin{equation}
    H(\textbf{J}_{i,n}) = - \sum_{l,h} \bar{g}_{l,h} log(\bar{g}_{l,h}),
\end{equation}

where $\bar{g}_{l,h}$ is the generic element of the normalized version of the adjacency matrix $\textbf{J}_{i,n}$. This quantity decreases as the graph increases its homogeneity in the distances between reference instances and in the discrepancy of the relative semantics. The entropy also increases proportionally to the occurrences of neutral semantics, as stimuli eliciting neutral emotions were considered as non-polarizing towards positivity or negativity.

The defined energy is a measure of how much the artificial agent is “conscious” of stimuli associated with positive semantics, i.e., a measure of how the agent's experiences are polarized towards positivity. The defined entropy, instead, is a measure of how much the artificial agent is “conscious” of stimuli heterogeneously associated with different semantics, i.e., a measure of how agent experiences are not polarized towards a specific semantic.

\section{Experiments with visual stimuli}
\label{sec:experimental_results}

Cognitive layers of Attention, Awareness, and Consciousness were tested together with the Smart Sensing, i.e., with the cognitive layers of Sensation, Perception, Emotion, and Affection. The custom dataset employed for the present experimentation is the same adopted in the prodromal study. The Smart Sensing was re-executed by deploying, instead of XGBoost, a Fully Connected Neural Network (FCNN) as the learner, coupled with the VGG16 \cite{simonyan2014very} convolutional network for feature extraction. A total of 611 and 92 images were considered for training and testing the proposed model of computational consciousness. Regarding the Emotion cognitive layer, the new learner reached 89.13\% accuracy by considering the classes of anger, fear, happiness, surprise, contempt, sadness, disgust, and the neutral state, while 94.57\% accuracy by considering the positive, negative, and neutral classes only (surprise and happiness were considered as belonging to the set of positive emotions).

Table \ref{table:confusion_matrix_emotion} summarizes the results achieved by deploying the FCNN as the learner for generating the emotional spectrum $\textbf{E}_{s}$ at each time instant $n$. The network consisted of four dense layers of 64, 32, 16, and 8 units with ReLU activation, coupled with dropouts set to a percentage of 30\%. Optimization was performed through Adam at the learning rate of $10^{-3}$, with a batch size of 64 samples per epoch. The percentages of emotions in the considered dataset are: 11\% for happiness, 11\% for anger, 18\% for fear, 11\% for surprise, 6\% for contempt, 6\% for sadness, 16\% for disgust, and 21\% for the neutral state.

\begin{table}
 \caption{Performance in classifying Emotion cognitive instances through FCNN.}
  \centering
  \begin{tabular}{ccccccccc}
    \toprule
    \multicolumn{9}{c}{Confusion matrix of Emotion cognitive instances classification}                                 \\
    \midrule
              & Neutral & Happiness & Anger & Fear & Surprise & Contempt & Sadness & Disgust                           \\
    \midrule
    Neutral   & 18      & 0         & 0     & 0    & 0        & 0        & 0       & 0                                 \\
    Happiness & 1       & 7         & 0     & 0    & 0        & 0        & 1       & 0                                 \\
    Anger     & 0       & 0         & 10    & 1    & 0        & 0        & 0       & 0                                 \\
    Fear      & 1       & 0         & 0     & 17   & 1        & 0        & 0       & 0                                 \\
    Surprise  & 0       & 0         & 0     & 0    & 7        & 0        & 0       & 2                                 \\
    Contempt  & 0       & 0         & 0     & 1    & 0        & 6        & 0       & 0                                 \\
    Sadness   & 0       & 0         & 0     & 0    & 0        & 2        & 5       & 0                                 \\
    Disgust   & 0       & 0         & 0     & 0    & 0        & 0        & 0       & 12                                \\
    \bottomrule
  \end{tabular}
  \label{table:confusion_matrix_emotion}
\end{table}

Emotion cognitive instances were well classified, except for what regards the emotions of surprise and sadness, which were misclassified four times. The suitability of the performance allowed us to test Attention, Awareness, and Consciousness with better performance of the Smart Sensing (4.13\% more accuracy, compared to the prodromal study). 

Table \ref{table:uncertainty_info_incompleteness} shows the values adopted for the plausibility, credibility, and possibility scores (i.e., the support in info-incompleteness) associated with the stimuli provided to the model. The values were chosen randomly, as a multi-agent experimentation of the proposed model was not intended as an objective of the present study. Even though a simulation of the model with support info-incompleteness was performed, the present study focuses on the definition and experimentation of a model of computational consciousness for non-interacting agents.

\begin{table}
 \caption{The plausibility, credibility, and possibility scores adopted for each stimulus in the experimentation concerning the support in info-incompleteness.}
  \centering
  \begin{tabular}{cccc}
    \toprule
    \multicolumn{4}{c}{Adopted scores for the support in info-incompleteness conditions} \\
    \midrule
    Stimulus                     & Plausibility ($Pl(\textbf{r}_{i,n})$)      & Credibility ($Cr(\textbf{r}_{i,n})$)     & Possibility ($Po(\textbf{r}_{i,n})$)     \\
    \midrule
    Beautiful woman              & 0.2               & 0.6             & 0.4             \\
    Own home                     & 0.9               & 0.8             & 0.5             \\
    Murder of animal             & 0.8               & 0.6             & 0.8             \\
    War                          & 0.3               & 0.9             & 0.2             \\
    Man pointing weapon          & 0.6               & 0.8             & 0.5             \\
    Terrorism                    & 0.1               & 0.8             & 0.5             \\
    Crazy sportsman              & 0.7               & 0.2             & 0.9             \\
    Politician                   & 0.9               & 0.9             & 0.9             \\
    Parking car                  & 0.9               & 0.8             & 0.9             \\
    Someone's death or sick      & 0.2               & 0.4             & 0.6             \\
    Car accident                 & 0.5               & 0.8             & 0.7             \\
    Injury                       & 0.9               & 0.7             & 0.4             \\
    Autopsy                      & 0.1               & 0.3             & 0.6             \\
    Landscape                    & 0.9               & 0.9             & 0.8         
    \\
    \bottomrule
  \end{tabular}
  \label{table:uncertainty_info_incompleteness}
\end{table}

Significance testing concerning the achieved results was performed through one-way ANOVA, of which 95\% confidence intervals, specified through mean and standard deviation (i.e., $M$ ($SD$)), are shown in the tables; salient evidence is highlighted in the text by specifying the $F$-score and $p$-value.

The Attention, Awareness, and Consciousness cognitive layers were experimented both on the training and test sets of emotional episodes. The average of the components in cognitive instances $\textbf{r}_{5,n}$  and the percentage of Attention obtained for each stimulus and emotional class were computed. 

\begin{figure}[!h]
    \centering
    \includegraphics[width=\textwidth]{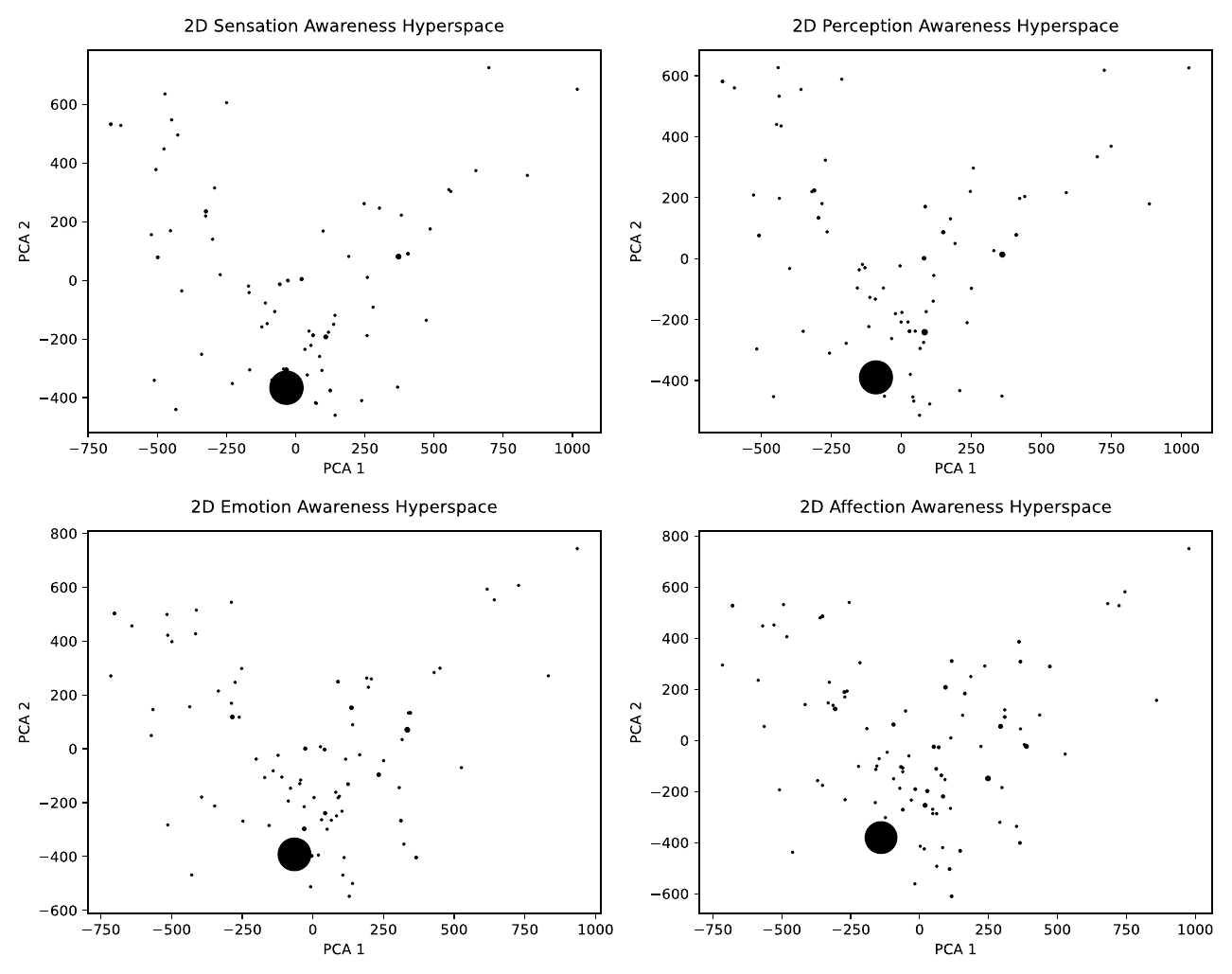}
    \caption{Reference instances $\textbf{r}_{1,m}$, $\textbf{r}_{2,m}$, $\textbf{r}_{3,m}$, $\textbf{r}_{4,m}$, belonging to the sets $M_{1,n-1}$, $M_{2,n-1}$, $M_{3,n-1}$, $M_{4,n-1}$, associated with the Sensation, Perception, Emotion, and Affection cognitive layers, respectively, in their related Awareness Hyperspaces. The cognitive instances are plotted with size proportional to the $Pr(\textbf{r}_{i,m})$ values.}\label{fig:awareness_hyperspaces_sens_perc_emot_affe}
\end{figure}

The Awareness cognitive layer was experimented with optimal $R = 2007$ both on the training and test sets regarding the provided emotional episodes. Figure \ref{fig:awareness_hyperspaces_sens_perc_emot_affe} shows a visual representation of the reference cognitive instances $\textbf{r}_{1,m}$, $\textbf{r}_{2,m}$, $\textbf{r}_{3,m}$, $\textbf{r}_{4,m}$, belonging to the sets $M_{1, n-1}$, $M_{2, n-1}$, $M_{3, n-1}$, $M_{4, n-1}$, in the Sensation, Perception, Emotion, and Affection Awareness Hyperspaces, respectively. To provide a 2D representation of the results, the two principal components describing the most of the variance were generated by employing PCA (Principal Component Analysis) dimensionality reduction \cite{jolliffe2005principal}. Each point in the Figure represents a reference instance $\textbf{r}_{i,m}$, which relative probability was highlighted through the size of the points themselves. For instance, the biggest point shown in the plots is associated with the Awareness cognitive instance concerning the “landscape” stimulus.

Regarding Consciousness, social and personal morality instances were computed by evaluating the emotional classification of the learner to positive, negative, and neutral classes. In particular, Sensation, Perception, Emotion, and Affection cognitive instances acquired at a time instant in which the learner provided the emotions of anger, fear, contempt, sadness, and disgust are associated with a negative semantic, while associated with a positive semantic for the emotions of happiness and surprise and with a neutral semantic for the neutral state. In principle, social and personal morality instances should be different, since a subject can express disagreement in the semantics related to cognitive instances; for instance, a terrorist can subjectively associate the “terrorism” stimulus with happiness, which, under the assumptions considered in the present study, is associated with a positive semantic, while being aware that for the society the same stimulus is associated with the emotion of fear, which, under the same aforementioned assumptions, is associated with a negative semantic. Instead, the present experimentation considers artificial agents which agree on the Environment and Subjective semantics.

\begin{figure}[!h]
    \centering
    \includegraphics[width=0.9\textwidth]{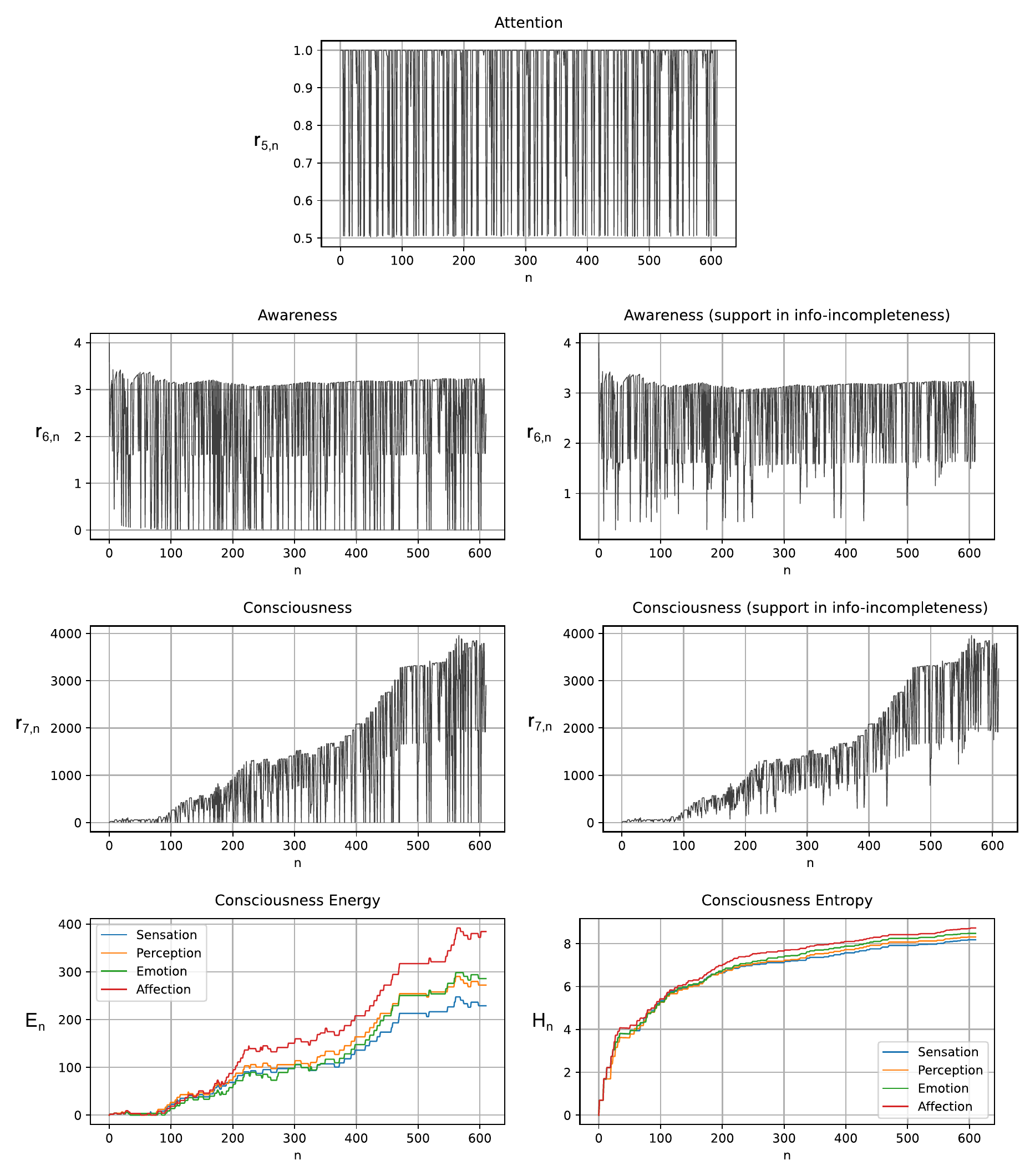}
    \caption{Trends of cognitive instances $\textbf{r}_{5,m}$, $\textbf{r}_{6,m}$, $\textbf{r}_{7,m}$, associated with the Attention, Awareness, and Consciousness cognitive layers, respectively, together with the related trends of energy and entropy. The plots concerning Awareness and Consciousness are compared with their version involving support in info-incompleteness.}\label{fig:plots_recap}
\end{figure}

Figure \ref{fig:plots_recap} shows the plots concerning cognitive instances of Attention, Awareness, and Consciousness, together with the related energy and entropy, in the function of the time instants associated with the provided visual stimuli for the experiments. The Attention results do not change with support in info-incompleteness, while Awareness and Consciousness trends result more positive when considering the plausibility, credibility, and possibility scores associated with the stimuli. As it can be seen from the Figure, Awareness, and Consciousness upper bounds do not change, while their lower bounds significantly increase, when providing support in info-incompleteness to the artificial agent. Significant higher Awareness and Consciousness are achieved for poorly experienced cognitive instances, i.e., instances characterized by low probability $Pr(\textbf{r}_{i,n})$, when considering plausibility, credibility, and possibility scores. 

It is also noticeable that the initial Awareness cognitive instance provided a maximum peak. This state indicates the moment of agent's activation, i.e., the acquisition of the first Awareness cognitive instance. The proposed model provided maximum probability to the first stimulus received since the ratio between experience and lifetime reached an upper bound in that instant. The Awareness cognitive layer with support in info-incompleteness presented significant higher magnitude since the uncertainty concerning the acquired cognitive instances decreased substantially.

Tables \ref{table:statistics_no_info_incompleteness} and \ref{table:statistics_info_incompleteness} show the statistics of the experiments concerning the proposed model of computational consciousness performed without and with support in info-incompleteness.

\begin{table}[h!]
\centering
\caption{Statistics concerning experiments involving no support in info-incompleteness.}
\begin{tabular}{cccc}
\toprule
\multicolumn{4}{c}{Experiments’ statistics (no support in info-incompleteness)}             \\
\midrule
\multicolumn{4}{c}{CIs of Attention, Awareness, and Consciousness for each emotional class} \\
\midrule
Stimulus                      & Attention ($\textbf{r}_{5,n}$)        & Awareness ($\textbf{r}_{6,n}$)        & Consciousness ($\textbf{r}_{7,n}$)         \\
\midrule
Neutral                       & 1.00 (0.00)      & 3.12 (0.07)      & 1621.07 (218.71)      \\
Happiness                     & 0.99 (0.01)      & 2.43 (0.32)      & 1349.92 (310.43)      \\
Anger                         & 0.99 (0.01)      & 2.02 (0.30)      & 1334.17 (308.65)      \\
Fear                          & 0.98 (0.01)      & 2.69 (0.21)      & 1391.34 (246.13)      \\
Surprise                      & 0.99 (0.01)      & 2.01 (0.32)      & 922.63 (255.16)       \\
Contempt                      & 0.99 (0.01)      & 2.78 (0.34)      & 1636.38 (451.75)      \\
Sadness                       & 0.51 (0.00)      & 1.46 (0.15)      & 835.06 (234.18)       \\
Disgust                       & 0.51 (0.00)      & 1.33 (0.11)      & 674.83 (130.65)       \\
\midrule
\multicolumn{4}{c}{CIs of Attention, Awareness, and Consciousness for each stimulus}        \\
\midrule
Stimulus                      & Attention ($\textbf{r}_{5,n}$)        & Awareness ($\textbf{r}_{6,n}$)        & Consciousness ($\textbf{r}_{7,n}$)         \\
\midrule
Parking car                   & 0.97 (0.04)      & 3.08 (0.12)      & 1710.26 (699.78)      \\
Man pointing weapon           & 0.99 (0.01)      & 2.44 (0.61)      & 1083.63 (559.24)      \\
Someone’s death or sick       & 0.57 (0.07)      & 1.76 (0.21)      & 1304.62 (357.60)      \\
Autopsy                       & 0.52 (0.02)      & 1.40 (0.14)      & 732.16 (159.09)       \\
Injury                        & 0.51 (0.00)      & 1.29 (0.25)      & 738.83 (329.58)       \\
Car accident                  & 0.58 (0.13)      & 1.26 (0.81)      & 297.52 (316.78)       \\
Terrorism                     & 0.98 (0.01)      & 2.78 (0.26)      & 1805.56 (412.10)      \\
Beautiful woman               & 1.00 (0.00)      & 2.33 (0.36)      & 1125.61 (302.58)      \\
War                           & 1.00 (0.01)      & 2.83 (0.30)      & 1448.59 (340.52)      \\
Politician                    & 1.00 (0.01)      & 2.53 (0.57)      & 1737.19 (639.30)      \\
Landscape                     & 0.99 (0.01)      & 3.09 (0.08)      & 1634.10 (218.24)      \\
Murder of animal              & 0.99 (0.01)      & 1.75 (0.36)      & 979.75 (315.08)       \\
Own home                      & 1.00 (0.00)      & 3.18 (0.03)      & 2113.64 (907.05)      \\
Crazy sportsman               & 1.00 (0.00)      & 1.81 (0.36)      & 729.98 (234.38)       \\
\bottomrule
\end{tabular}
\label{table:statistics_no_info_incompleteness}
\end{table}

\begin{table}[h!]
\centering
\caption{Statistics concerning experiments involving support in info-incompleteness.}
\begin{tabular}{cccc}
\toprule
\multicolumn{4}{c}{Experiments’ statistics (support in info-incompleteness)}                \\
\midrule
\multicolumn{4}{c}{CIs of Attention, Awareness, and Consciousness for each emotional class} \\
\midrule
Stimulus                      & Attention ($\textbf{r}_{5,n}$)        & Awareness ($\textbf{r}_{6,n}$)        & Consciousness ($\textbf{r}_{7,n}$)         \\
\midrule
Neutral                       & 1.00 (0.00)       & 3.18 (0.02)      & 1648.65 (216.64)     \\
Happiness                     & 0.99 (0.01)       & 2.58 (0.25)      & 1407.96 (303.01)     \\
Anger                         & 0.99 (0.01)       & 2.71 (0.13)      & 1714.93 (270.05)     \\
Fear                          & 0.98 (0.01)       & 2.87 (0.13)      & 1493.18 (233.96)     \\
Surprise                      & 0.99 (0.01)       & 2.56 (0.16)      & 1270.04 (236.94)     \\
Contempt                      & 0.99 (0.01)       & 3.04 (0.11)      & 1815.69 (429.59)     \\
Sadness                       & 0.51 (0.00)       & 1.53 (0.09)      & 855.59 (233.31)      \\
Disgust                       & 0.51 (0.00)       & 1.45 (0.07)      & 710.58 (126.20)      \\
\midrule
\multicolumn{4}{c}{CIs of Attention, Awareness, and Consciousness for each stimulus}        \\
\midrule
Stimulus                      & Attention ($\textbf{r}_{5,n}$)        & Awareness ($\textbf{r}_{6,n}$)        & Consciousness ($\textbf{r}_{7,n}$)         \\
\midrule
Parking car                   & 0.97 (0.04)       & 3.08 (0.12)      & 1710.26 (699.78)     \\
Man pointing weapon           & 0.99 (0.01)       & 2.84 (0.27)      & 1263.25 (518.29)     \\
Someone’s death or sick       & 0.57 (0.07)       & 1.78 (0.21)      & 1331.66 (366.26)     \\
Autopsy                       & 0.52 (0.02)       & 1.47 (0.11)      & 764.44 (154.03)      \\
Injury                        & 0.51 (0.00)       & 1.50 (0.10)      & 769.57 (323.26)      \\
Car accident                  & 0.58 (0.13)       & 1.56 (0.54)      & 383.81 (308.66)      \\
Terrorism                     & 0.98 (0.01)       & 2.90 (0.17)      & 1928.73 (389.03)     \\
Beautiful woman               & 1.00 (0.00)       & 2.50 (0.29)      & 1186.06 (297.64)     \\
War                           & 1.00 (0.01)       & 2.94 (0.20)      & 1490.57 (326.12)     \\
Politician                    & 1.00 (0.01)       & 2.98 (0.19)      & 2046.74 (568.30)     \\
Landscape                     & 0.99 (0.01)       & 3.14 (0.04)      & 1660.33 (216.32)     \\
Murder of animal              & 0.99 (0.01)       & 2.61 (0.15)      & 1441.64 (280.87)     \\
Own home                      & 1.00 (0.00)       & 3.18 (0.03)      & 2113.64 (907.05)     \\
Crazy sportsman               & 1.00 (0.00)       & 2.48 (0.19)      & 1145.50 (223.75)   
  \\
\bottomrule

\end{tabular}
\label{table:statistics_info_incompleteness}
\end{table}

The results show that the Attention cognitive instances assume the highest magnitude for the stimuli related to the emotions of happiness, anger, fear, surprise, contempt, and the neutral state, while a decrease was found concerning those associated with sadness and disgust. The stimuli for which the artificial agent provided significant high Attention are “beautiful woman,” “own home,” and “crazy sportsman” ($F(13,597) = 434.16$, $p < 0.001$), which all belong to the set of positive emotions. The stimuli for which the artificial agent provided significant low Attention are “autopsy” and “injury” ($F(13,597) = 434.16$, $p < 0.001$), belonging to the set of disgust, while “someone's death or sick,” and “car accident” ($F(13,597) = 434.16$, $p < 0.001$), belonging to the set of sadness. The emotion for which the agent provided the highest Attention was happiness ($F(13,597) = 3101.39$, $p < 0.001$); contrarily, the most significant decrease was found during disgust and sadness ($F(13,597) = 3101.39$, $p < 0.001$).

In the experiments conducted without support in info-incompleteness, the Awareness and Consciousness cognitive instances are proportional to the probability score, i.e., to the frequency of the related stimuli. In particular, anger and surprise, which were characterized by the same percentage of occurrences, did not provide significant differences in the Awareness. However, the Awareness also depends on the number of cognitive instances encapsulated into hyperspheres and the number of hyperspheres associated with a given stimulus. Significant high Awareness was found for the emotion of contempt, which was characterized by the lowest percentage of occurrences ($F(13,597) = 31.87$, $p < 0.001$). Compared to the stimuli associated with the emotion of disgust, which were significantly more frequent in the occurrences, the stimuli related to the contempt resulted in a lower number of hyperspheres. The stimuli associated with the contempt presented less visual variance compared with those eliciting disgust. The artificial agent provided the highest and lowest Awareness during the neutral state and the emotion of disgust ($F(13,597) = 31.87$, $p < 0.001$), respectively; the stimuli for which the agent provided the highest and lowest Awareness are “own home” and “injury,” respectively ($F(13,597) = 20.10$, $p < 0.001$). Regarding Consciousness, the artificial agent provided the highest magnitude for the emotion of contempt and the neutral state (for which no significant differences were found with each other). The lowest Consciousness, instead, was found during the emotion of disgust ($F(13,597) = 7.46$, $p < 0.001$). Regarding the stimuli eliciting happiness and anger, no differences in the Consciousness cognitive instances were found. The stimuli for which the agent provided the highest and lowest Consciousness are “own home” and “car accident,” respectively ($F(13,597) = 6.50$, $p < 0.001$). Regarding the same cognitive layer, no significant differences were found concerning the stimuli of “autopsy,” “injury,” and “crazy sportsman;” the same holds for “parking car” and “politician.”

Regarding Consciousness entropy, the Sensation cognitive layer provided lower magnitude compared with Emotion ($F(1,609) = 3.86$, $p = 0.04$), while Perception was found to be comparable with Sensation and Emotion. The Affection cognitive layer, instead, provided significant high entropy ($F(3,607) = 8.13$, $p < 0.001$). Significant low Consciousness energy was found concerning the Sensation ($F(3,607) = 36.59$, $p < 0.001$), while no significant differences were found in the Perception and Emotion cognitive layers. Finally, the Affection provided the highest Consciousness energy.

\section{Discussion}
\label{sec:discussion}

The results concerning Attention are concordant with the scientific evidence regarding covert attention in human beings. In particular, a decrease in the attention associated with the emotion of disgust was found by Van Hooff et al. \cite{van2013disgust}, which investigated the effects of disgust-, fear-, and neutral-related stimuli on covert attention. They proposed an experiment, called \textit{covert orienting task}, by employing the IAPS (International Affective Picture System) dataset \cite{lang2005international} for studying participants' reactions to images. The process consisted of the task of targeting a picture, shown on a screen, eliciting one of the above emotions (i.e., disgust, fear, or the neutral state) in participants. The results provided significantly less accurate identifications and longer reaction times in targeting disgust-evoking pictures. An extension of the above study \cite{van2014disgust} revealed a significant lowering in the attention concerning disgust-evoking pictures compared to the emotion of happiness. No further studies were found concerning the same experimental method applied to the emotions of surprise, contempt, and sadness. Experimentation employing the covert orienting task for analyzing attention during the elicitation of the above emotions could confirm the relative results obtained in the present work. For instance, finding a lowering in the covert attention during the emotion of sadness would provide further evidence that the proposed attention model consistently describes the phenomenon. Adopting another experimental method, called the \textit{flanker task}, Bellaera and von Mühlenen \cite{bellaera2017effect} found that participants showed significant narrowing of the selective attention in conditions of sadness. Selective attention may be considered as the opposite of covert attention; thus, an experiment employing the covert orienting task may provide evidence for the decrease in the identification accuracy and longer reaction times. Research on the covert and selective attention often present conflicting results: Finucane \cite{finucane2011effect} proved that response times are significantly faster in anger and fear conditions than in the neutral state by employing the flanker task, while Van Hooff et al. \cite{van2013disgust} found no significant differences in the covert attention for the emotions of anger, fear, and the neutral state. 

A comparison of the Awareness cognitive layer with the scientific evidence is difficult. However, the results seem coherent with the evidence provided by Atas et al. \cite{atas2013repeating}, in which a gradual increase in the awareness of a visual stimulus was found by augmenting the repetitions. The experiment conducted by the authors was the \textit {visual masking} \cite{breitmeyer2006visual}, which consists in proposing the repetition of masked visual stimuli to participants. Visual masking represents one of the most employed methods for the study of consciousness, especially as regards the differentiation between conscious and unconscious processes. To our knowledge, no experiments concerning the study of awareness in the function of the experiences of other subjects have been conducted yet. However, the modeling of the cognitive differentiation between one's personal experience and that of other subjects, i.e., the evaluation of the probability score as concurrent with the plausibility, credibility, and possibility scores introduced the element of \textit{self-awareness} in the Awareness cognitive layer. Scientific evidence considers the \textit{mirror test} as one of the most relevant experimental methods for identifying self-awareness in species. Regarding the human being, the experimentation consists in verifying the capability of distinguishing one's image reflected in the mirror to the visualization of other targets. Since the above test was passed by humans, the proposed model of computational consciousness may introduce interesting perspectives of research, since the differentiation of probability, plausibility, credibility, and possibility scores may reveal, from the “experiential” point of view, a sort of self-awareness in the machine. 

The experiments revealed that the probability associated with a given reference cognitive instance represents a measure of the “quantity” of Awareness, intended as “artificial experience,” concerning a given event. In the proposed model, the magnitude of the Awareness cognitive instance $\textbf{r}_{6,n}$ increases proportionally to the number of occurrences acquired at a distance lower than or equal to $R$ from the relative reference instance $\textbf{r}_{i,m}$ (i.e., as the population of cognitive instances grows within a hypersphere). As defined in Section 5.2, the Awareness cognitive layer also integrates the contribution provided by non-direct experience. The awareness of a subject concerning a given stimulus was assumed as affected also by the interaction with other subjects; in fact, the acquisition of information associated with a given event can significantly influence the awareness of a subject. For instance, an external opinion regarding the content of a bi-stable image can be essential to the awareness of that visual stimulus. In the present model, while the probability-based model allows increasing the magnitude of Awareness cognitive instances directly, the characterization based on plausibility, credibility, and possibility was theoretically associated with the results obtained from external models or humans. In the case the agent reached not suitable $Pr(\textbf{r}_{i,n})$, the values $Pl(\textbf{r}_{i,n})$, $Cr(\textbf{r}_{i,n})$, and $Po(\textbf{r}_{i,n})$ can be acquired from external sources. For instance, these scores can be obtained by interacting with another agent, with a human, or by extracting data from the web.

The experiments revealed that Consciousness is higher during the processing of stimuli eliciting the emotions of contempt and the neutral state, while lower during disgust. The results highlight that the agent's experiences are more polarized towards positive semantics during contempt and the neutral state, while towards negative semantics during disgust. The model valued the comparison with positive experiences during the processing of stimuli such as “politician,” “parking car,” and “landscape,” while significantly comparing “autopsy” and “injury” stimuli with negative experiences. This process of comparison is affected by the contribution of the Consciousness energy, which provides a metric of semantic polarization in the Consciousness Graphs. By providing visual stimuli to participants, Schnall et al. \cite{schnall2008disgust} provided evidence for increasing judgments of wrongness for occurrences eliciting disgust, compared with those eliciting sadness. Similarly, the present study found a lower Consciousness magnitude during sadness than disgust, revealing less negative orientation of personal morality instances towards stimuli eliciting sadness, such as “someone's death or sick” and “car accident.” However, the authors also found that participants provided less severe judgments towards visual stimuli concerning sadness, compared with those concerning the neutral state. This last result was found as opposed to the outcomes of our experiments, in which Consciousness energy revealed higher positiveness during the processing of stimuli related to the neutral state. 

The analysis of entropy and energy revealed that Consciousness increases as cognition approaches deeper cognitive layers. Sensation Consciousness was found to be weaker than Affection Consciousness, in terms of entropy and energy. The results reveal that the experimented artificial agent acquires higher Consciousness for the personal, as opposed to the social, morality since the Affection cognitive layer provided the increasing positive orientation of personal morality instances. These results encourage a scientific investigation of consciousness in the function of sensations, perceptions, emotions, and affections through the metrics of entropy and energy.

In general, the proposed model of computational consciousness approximates several human cognitive tasks, such as covert attention, the awareness related to a repetition of stimuli through both direct and indirect experiences, and consciousness as the semantic connection of environment and subjective experiences. The mechanisms conceived in Section \ref{sec:consciousness_background} as “objective” and “phenomenal,” in our model were represented by the Environment and Subjective Awareness; those conceived as “subjective” and “access,” in the proposed model were represented by the Environment and Subjective Consciousness.

\section{Conclusions and future work}
\label{sec:conclusions}

In the present study, a novel model of computational consciousness for non-interacting agents was proposed. By analyzing the latest advancements in the field of AC and the Smart Sensing prodromal solution, an info-structural hierarchy of cognitive layers concerning the cognitive tasks of sensation, perception, emotion, affection, attention, awareness, and consciousness was defined and experimented on visual stimuli. In doing so, the theoretical background regarding the studies of consciousness was evaluated and several theories, definitions, and scientific evidence were considered for operating the mathematical modeling. Consciousness was assumed as the moral evaluation of environment and subjective experiences, which depend on the emotional activity.

Attention was deployed computationally through a bottom-up approach and adopted as the fundamental modulating factor for awareness to happen; its mathematical representation was conceived with a matrix acquiring artificial sensation, perception, emotion, and affection stimuli. Emotions were represented through the classes of anger, fear, happiness, surprise, contempt, disgust, and the neutral state. Adopting the theory of decision and reasoning in uncertainty and info-incompleteness conditions, Awareness was modeled as the information fusion of probability, plausibility, credibility, and possibility scores concerning the acquired sensation, perception, emotion, and affection related artificial experiences. In our solution, the probability was deployed to represent the agent's direct experience of stimuli, while plausibility, credibility, and possibility represent the agent's indirect experience. Our model of awareness was supported by the geometry of hyperspheres, which allow representing information in the Euclidean space. Finally, consciousness was modeled by connecting and associating the above artificial experiences with semantics related to the sets of social and personal morality employing graph structures. The final result consisted of an index obtained by computing the sum of contributions related to the analysis of the above graph structures in terms of energy and entropy.

The results showed that, from our comparative analysis, the proposed model of attention is concordant with the scientific evidence concerning covert attention in human beings. Regarding our models of awareness and consciousness, a direct comparison with scientific studies was difficult, but strong similarities were found concerning experiments involving the repetition of visual stimuli and the relative moral judgments. In particular, the proposed model resulted comparable with scientific evidence regarding increasing judgments of wrongness concerning the occurrences of visual stimuli eliciting disgust, compared with those eliciting sadness.

The present model of computational consciousness provides contributions to AC by proposing an approach to implement artificial subjectivity in the machine, as well as by investigating the sequential, rather than the parallel, execution of cognitive tasks. Furthermore, a model enhancing the functioning of consciousness in terms of emotional activity and morality has been defined and experimented with visual stimuli. The unique part of the system involving the training of a learner is the model deployed for artificial emotion, which allows running the solution on specific emotional history. From a scientific point of view, the results obtained in the present study encourage investigations on the subjective nature of consciousness through emotional, moral, energetic, and entropic perspectives.

A future study concerning the proposed model consists in the introduction of causality by associating the arcs of Consciousness Graphs to the actions performed by the agent towards the environment. This approach permits to adopt the paradigm of reinforcement learning to find the optimal policy for reaching the best reward in never-explored environments. Such a model could be deployed in social robots with the aim of studying computational consciousness in human-robot interaction. A further future study regards the acquisition of plausibility, credibility, and possibility scores for the proposed model of awareness directly from the interacting environment or from other sources of information, such as the internet.

\bibliographystyle{unsrt}  
\bibliography{templateArxiv}

\end{document}